\documentclass[letterpaper, 10 pt, conference]{ieeeconf}  %

\IEEEoverridecommandlockouts                              %

\usepackage{graphicx} %
\usepackage[noadjust]{cite}
\usepackage{multirow}
\usepackage{amsfonts}
\usepackage{amsmath}
\usepackage[ruled,vlined,lined,linesnumbered,boxed,commentsnumbered]{algorithm2e}

\newcommand{\Expectof}[1]{\mathbb{E} \left[ #1 \right] }

\title{\LARGE \bf
Good Feature Selection for Least Squares Pose Optimization in VO/VSLAM
}

\author{Yipu Zhao$^{1}$ and Patricio A. Vela$^{1}$%
\thanks{$^{1}$Yipu Zhao {\tt\small yipu.zhao@gatech.edu} and Patricio A. Vela {\tt\small pvela@gatech.edu} are with Department of Electrical and Computer Engineering, Georgia Institute of Technology, Atlanta, Georgia, USA. }
\thanks{This work was supported, in part, by the National Science Foundation under Grant No. 1400256 and 1544857.}%
}

\usepackage{mymacros}

\graphicspath{ {./images/}}

\begin{document}

\maketitle
\thispagestyle{empty}
\pagestyle{empty}

\begin{abstract}

This paper aims to select features that contribute most to the pose
estimation in VO/VSLAM.  Unlike existing feature selection works that are
focused on efficiency only, our method significantly improves the accuracy 
of pose tracking, while introducing little overhead.  By studying the impact of 
feature selection towards least squares pose optimization, we
demonstrate the applicability of improving accuracy via good feature
selection.  To that end, we introduce the {\em Max-logDet} metric to
guide the feature selection, which is connected to the conditioning of 
least squares pose optimization problem.  We then describe an efficient
algorithm for approximately solving the NP-hard {\em Max-logDet} problem.  
Integrating {\em Max-logDet} feature selection into a state-of-the-art
visual SLAM system leads to accuracy improvements with low overhead,
as demonstrated via evaluation on a public benchmark.

\end{abstract}

\section{Introduction}

Least squares optimization techniques, such as Gauss-Newton and
Levenberg-Marquardt methods, are widely used for optimizing camera pose
in state-of-the-art VO/VLAM systems for robotics
(e.g. ORB-SLAM\cite{ORBSLAM}, SVO\cite{SVO2017}, DSO\cite{DSO2017}).
Unfortunately, least squares are sensitive to perturbations in the
source data.  Incorporating robust influence functions mitigates this
problem, but does not completely suppress the induced error.
In VO/VSLAM, the perturbations from both measurements (e.g.
noisy features/patches) and references (e.g. inaccurate mapping) 
negatively affects pose optimization with least squares techniques.
Regarding accurate pose optimization, not all features/patches
being matched contribute the same.  If only those valuable towards
accurate pose estimation are utilized, the total amount of noise
introduced into the least squares can be reduced, while preserving the
conditioning of the optimization problem.

\begin{figure*}[!htb]
\centering
\includegraphics[width=0.8\linewidth]{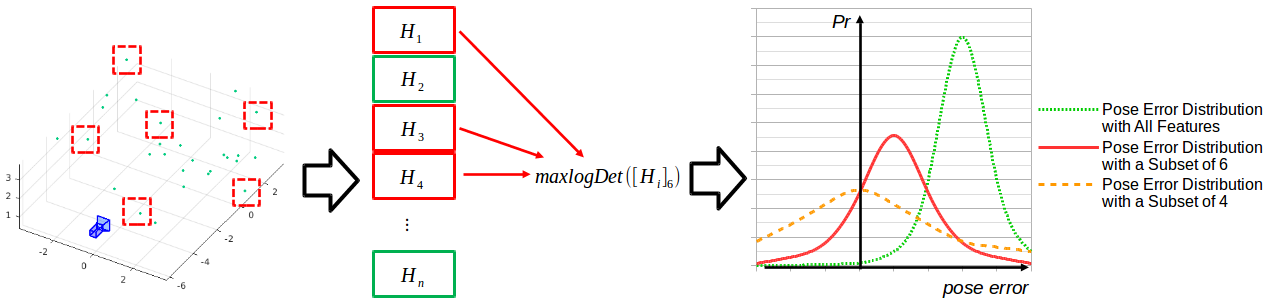}
\vspace*{-0.10in}
\caption{A toy case to illustrate our approach.  For least squares pose optimization, feature selection is equivalant to row block selection of Jacobian.  With the {\em Max-logDet} metric, the subset of row blocks (and the corresponding feature subset) can be obtained.  With a properly sized feature subset (e.g. the error distribution in red solid line), the bias of pose optmization is significantly reduced, while an acceptable amount of variance is introduced. \label{fig:Illustration}} 	
\vspace*{-0.15in}
\end{figure*}

The idea of enhancing the performance of VO/VSLAM with feature selection
is not novel.  Conventionally, fully data-driven and randomized methods
such as RANSAC are used to reject outlier features \cite{MonoSLAM}.
Extensions to RANSAC improve its computational efficiency
\cite{KalmanSAC,1pRANSAC}.  These RANSAC-like approaches are utilized in
many VO/VSLAM systems \cite{MonoSLAM,PTAM,ORBSLAM}.  However, the scope
of this paper is on ``inlier selection'', which differs from outlier
rejection: outlier rejection aims to remove clearly wrong matches,
while ``inlier selection'' aims to identify valuable inlier matches
from useless ones.  The two aspects are complementary. A high-level
overview of inlier-selection in SLAM can be found in
\cite{cadena2016past}.  

Image appearance has been commonly used to guide inlier selection:
feature points with distinct color/texture patterns are more likely to get
matched correctly \cite{GoodFeaturesToTrack,sala2006landmark,shi2013feature}.  
However, these works solely rely on quantifying distinct appearance,
while the structural information of the 3D world and the camera motion
are ignored.  While appearance cues are important in feature selection,
the focus of this paper is on the latter properties: selecting features
based on structural and motion information.  These two complementary
approaches should be combined into a general feature selection
methodology.

To exploit structural and motion information, covariance-based feature
selection methods are studied.  The pose covariance matrix
1) contains both structural and motion information implicitly, and 
2) approximately represents the uncertainty ellipsoid of pose estimation.  
Based on pose covariance, different metrics were introduced to guide the
feature selection, 
such as information gain \cite{activeSearch,covarianceRecovery}, 
entropy \cite{zhang2005entropy}, 
trace \cite{lerner2007landmark} and 
covariance ratio \cite{cheein2009feature}.  
A potential issue is the pursuit of low uncertainty in estimation,
rather than accuracy.  These two objectives are not equivalent; 
an estimate can converge to a wrong pose with high confidence.  In
addition, the works above target efficiency of pose tracking; none of 
them explicitly target accuracy improvements via feature selection.

The works most related to this paper are 
\cite{zhang2015optimally, zhang2015good} and \cite{carlone2017attention}.
In \cite{zhang2015optimally, zhang2015good}, the connection between pose
tracking accuracy and observability conditioning of SLAM as a dynamic
system was studied.  The insight of their work being: the better
conditioned the SLAM system is, the more tolerant the pose estimator
will be to feature measurement error.  To that end, the minimum singular
value of observability matrix is used in to assess the observability
condition of the SLAM system.  Here, we employ a different metric, 
{\em Max-logDet}, and demonstrate its superiority to minimal singular value.  
Furthermore, we argue that bundle adjustment pipelines may benefit from
an alternative set of matrices to consider for solution conditioning,
ones more related to the underlying bundle adjustment problem.

In \cite{carlone2017attention}, feature selection is performed by
maximizing the information gain of pose estimation within a
prediction horizon.  Two feature selection metrics were evaluated, 
minimal eigenvalue and log determinant ({\em Max-logDet}). 
Though the current investigation uses the log determinant metric, the 
algorithm for approximately selecting the {\em logDet} maximizing feature 
subsetdiffers, as does the matrix whose conditioning is optimized.  
We propose a lazier-greedy algorithm taking an order of magnitude less
time than the greedy algorithm of \cite{carlone2017attention}, yet
preserving the optimality bound.
Further, we are interested in improving the accuracy of pose tracking by
selecting features with robustness properties, while preserving the time
cost.  As illustrated in Fig~\ref{fig:Illustration}, this objective can
be achieved by balancing between variance (e.g. minimizing the
uncertainty of pose estimation) and bias (e.g. minimizing the
expectation of pose error).

The proposed method hits all studied aspects to date: outlier rejection,
appearance-based inlier selection, and structural-based inlier
selection.  We show that all three together outperform each
individually.  The contributions are:

\noindent
1) Demonstrated \textbf{applicability} of improving accuracy via feature 
selection, mathematically and experimentally; \\
2) Exploration of \textbf{metrics} connected to the least squares
conditioning of pose optimization, with quantification of {\em
Max-logDet} as the optimal metric;  \\
3) An \textbf{efficient algorithm} to approximately solve the 
NP-hard {\em Max-logDet} problem for real-time feature selection in
the pose tracking step of VO/VSLAM; and \\
4) \textbf{Integration} of the algorithm \textbf{into a state-of-the-art
visual SLAM} system and evaluation on public benchmark collected with a
high-speed UAV.  By selecting good features with the proposed method, 
tracking accuracy is significantly improved with minimal impact on the
time cost.

\section{Feature Selection in Least Squares Pose Optimization}

The least squares objective of pose optimization in feature-based
VO/VSLAM can be written as follows,
\begin{equation}  \label{eq:LeastSquare}
  \arg \min \left\Vert h(x, p)-z \right\Vert^2,
\end{equation}
where $x$ is the pose of the camera, 
$p$ is the 3D feature points and 
$z$ the is corresponding measurements on 2D image frame.  
The measurement function, $h(x, p)$, is a combination of 
world-to-camera transformation and pin-hole 
projection.  We base the theory of feature selection upon this
objective function.  For direct VO/VSLAM, the objective function is
slightly different. Nevertheless, the theory in the following can be
easily extended to the direct version, once the direct residual term is
properly approximated to first-order.

Solving the least squares objective of Eq~(\ref{eq:LeastSquare}) often 
involves the first-order approximation to the non-linear measurement
function $h(x, p)$ linearization about initial guess $x^{(s)}$:
\begin{equation} \label{eq:FirstOrderApprox}
\left\Vert h(x, p) -z \right\Vert^2 
    = \left\Vert h(x^{(s)}, p) + H_x(x-x^{(s)}) - z \right\Vert^2.
\end{equation}
To minimize of the first-order approximation Eq~(\ref{eq:FirstOrderApprox})
via Gauss-Newton, the pose estimation is iteratively updated via
\begin{equation} \label{eq:GN_oneIter}
  x^{(s+1)} = x^{(s)} - H_x^+( z-h(x^{(s)}, p) ).
\end{equation}
Gauss-Newton accuracy is affected by two types of error: measurement 
error $\epsilon_z$ and map error $\epsilon_p$.  Again with the first-order 
approximation of $h(x, p)$ at the initial pose $x^{(s)}$ and assumed map
point $p^{(s)}$, we can connect the pose optimization error to measurement and
map errors:
\begin{equation} \label{eq:Error_Formula}
  \epsilon_x = H_x^+( \epsilon_z - H_p \epsilon_p ).
\end{equation}
Notice $H_p$ is a block diagonal matrix of size $2n\times 3n$, where $n$
is the number of matched features.  We will discuss the influence of
feature selection on pose optimization error $\epsilon_x$.

\paragraph{Minimizing the Variance from Measurement / Map Error}
Consider the case that only measurement error $\epsilon_z$ exists 
and is i.i.d. Gaussian with isotropic, diagonal covariance: 
$\epsilon_z(i) \sim N(0, \sigma_z^2)$.  
The pose covariance matrix will be 
\begin{equation} \label{eq:Pose_Cov_Meas}
  \Sigma_x = \sigma_z^2 (H_x^T H_x)^{-1}
           = \sigma_z^2 [ \sum_{i=1}^{n}{H_x(i)^T H_x(i)} ]^{-1}.
\end{equation}
where $H_x(i)$ being the corresponding row block in $H_x$ for feature $i$.
The pose covariance matrix represents the uncertainty ellipsoid in pose configuration 
space.  According to Eq~\eqref{eq:Pose_Cov_Meas}%
, one should always use all the 
features/measurements available to minimize the uncertainty (i.e. variance) in pose 
estimation: with more measurements, the singular values of the measurement Jacobian $H_x$ will increase 
in magnitude.  The worst case variance would be proportional to $\sigma_{min}^{-2}(H_x)$,
whereas in the best case it would be $\sigma_{max}^{-2}(H_x)$.

Similarly, consider minimizing the variance due to map error.  With an 
i.i.d. Gaussian assumption on map error: 
$\epsilon_p(i) \sim N(0, \sigma_p^2)$, we can derive the pose covariance
matrix,
\begin{equation} \label{eq:Pose_Cov_Map}
\begin{split}
  \Sigma_x & = \sigma_p^2 H_x^+ H_p H_p^T (H_x^+)^T \\
  		   & = \sigma_p^2 \{ \sum_{i=1}^{n}{H_x(i)^T [H_p(i) H_p(i)^T]^{-1} H_x(i)} \}^{-1}.
  \end{split}
\end{equation}
Still all map points being matched should be utilized.  
The worst case variance would be proportional to $\sigma_{min}^{-2}(H_p^+ H_x)$,
whereas in the best case it would be $\sigma_{max}^{-2}(H_p^+ H_x)$.

\paragraph{Minimizing the Bias from Map Error}
Yet another case to consider is the existence of biased map error 
(i.e. the mean of error distribution is non-zero).
Biased map error may appear in real VO/VSLAM applications.  
For example, the map points could be batch-perturbed when triangulated with erroneous camera poses.
Also, offset exists within a group of map points when they are jointly optimized with scale-drifted key frames.  
Here, we briefly discuss the case that map error $\epsilon_p$ 
follows non-zero-mean i.i.d. Gaussian, $\epsilon_p(i) \sim N(\mu_p, \sigma_p^2)$
and measurement error $\epsilon_z$ is unbiased.

The expectation of the pose optimization error will be biased by the 
non-zero-mean map error:
\begin{equation} \label{eq:Pose_Bias_Map}
  \Expectof {\epsilon_x}
    = \Expectof{H_x^+ H_p \epsilon_p}
    = H_x^+ H_p {\bf 1}_n \mu_p
\end{equation}
where ${\bf 1}_n$ is a tall matrix of $n$ smaller identity matrices.
In the worst case scenario, pose error expectation $\Expectof {\epsilon_x}$ is amplified by 
$\sigma_{max}(H_x^+ H_p)$, whereas in the best case it is only amplified 
by $\sigma_{min}(H_x^+ H_p)$.  
Subset selection affects the two components, $H_x$ and $H_p$, in
opposite ways: it will increase the amplification 
factor of $H_x^+$, while bounding the amount of noise induced by $H_p$.  
When the reduction of the latter is larger in magnitude than the
increase of the prior, the pose optimization error should drop.
Obviously, one possible objective of feature subset selection would 
be minimizing the factor of worst case scenario, $\sigma_{max}(H_x^+ H_p)$; 
another option would be minimizing both $\sigma_{max}(H_x^+ H_p)$ 
and $\sigma_{min}(H_x^+ H_p)$.

Furthermore, the two matrices, $H_x^+$ and $H_p$, can be combined into one.  
Move both to the left hand side of Eq~\eqref{eq:Pose_Bias_Map},
\begin{equation} \label{eq:Pose_Bias_Fin}
  H_p^+ H_x \Expectof {\epsilon_x}
    =  {\bf 1}_n \mu_p.
\end{equation}
Note that the projection Jacobian $H_p$ is a $2n\times 3n$ block
diagonal matrix, consisting of $2\times 3$ denoted by $H_p(i)$.
Meanwhile, each row block of $H_x$ can be written as $H_x(i)$.  

To remove the need for the pseudo inverse of $H_p$, add one more row 
$[0 \ 0 \ 1]$, to each block $H_p(i)$.  In addition a zero row is added
to each row block $H_x(i)$ to get new row block $H_x^n(i)$.
This trick does not affect the structure of the least square problem,
but it does allow inversion of the new diagonal block 
$H_p^n(i)=[H_p(i); 0 \ 0 \ 1]$.  
After performing block-wise multiplication, one can obtain
the combined matrix $H_c$, consisting of concatenated row blocks 
$H_p^n(i)^{-1} H_x^n(i)$.  Instead of working with two independent matrices 
$H_x$ and $H_p$, we consider optimizing the spectral properties
of their combination, $H_c$. %

This section covered three perspectives of pose optimization under
measurement \& map error, and identified the scenario whereby feature
selection might reduce estimation error.  Under biased map errors (which
is true in real VO/VSLAM applications), selecting a subset of features
could improve least squares pose optimization accuracy.

\section{Good Feature Selection Metrics}

Analyzing the impact of map error on least squares pose optimization 
led to equations where the singular values of $H_c$ and their
extremal properties were connected to best/worst case outcomes.  Actual
outcomes would depend on the overall spectral properties of $H_c$.
Therefore, we seek a sub-matrix of $H_c$ preserving as best as possible
the overall spectral properties, and at minimum the extremal spectral
properties.

Under this spectral preservation objective, the feature selection
problem is equivalent to selecting a subset of row blocks in the
matrix $H_c$ such that the norm of the selected sub-matrix is as large
as possible.  Once selected, the sub-matrix determines which
measurements from the set of available measurements should be taken
(these would be a subset of the good feature points).
Submatrix selection with spectral preservation has been extensively
studied in the fields of computational theory and machine learning
\cite{gu1996efficient,boutsidis2009improved}, for
which several matrix-revealing metrics exist to score the subset
selection process.  They are listed in Table~\ref{tab:Reveal-Metric}.

Subset selection with any of the matrix-revealing metrics listed above 
is equivalent to a finite combinational optimization problem:
\begin{equation} \label{eq:Combine_Opt}
  \max_{S\subseteq\{1,2,...,n\}, |S|=k} f([H_c(S)]^T [H_c(S)])
\end{equation}
where $S$ is the indices of selected row blocks from full matrix $H_c$, 
$[H_c(S)]$ is the corresponding concatenated submatrix, and 
$f$ is the matrix-revealing metric.

\begin{table}[h]
\centering
\caption{Commonly used matrix-revealing metrics} \label{tab:Reveal-Metric}
\begin{tabular}{ l | l }
{\em Max-Trace}				&	Trace $Tr(Q) = \sum_{1}^{k} Q_{ii}$ is max.	\\
{\em Min-Cond}				&	Condition $\kappa(Q) = \lambda_{1}(Q)/\lambda_{k}(Q)$ is min.	\\
{\em Max-MinEigenValue}		&	Min. eigenvalue $\lambda_{k}(Q)$ is max.	\\
{\em Max-logDet}				&	Log. of determinant $\log \det(Q)$ is max.	\\
\end{tabular}
\vspace*{-0.1in}
\end{table}

\subsection{Submodularity}
The combinational optimization above can be solved with brute-force, 
but the exponentially-growing problem space quickly becomes impractical 
to search, especially for real-time VO/VSLAM applications.  
Heuristics for subset selection target one structural property, 
{\em submodularity} \cite{summers2016submodularity,shamaiah2010greedy,carlone2017attention}.  
If a set function (e.g. matrix-revealing metric) is submodular and
monotone increasing, then approximate, greedy combinational optimization
of the set function (e.g. subset selection) has near optimality guarantees.  

Except for {\em Min-Cond}, all other three metrics list in 
Table~\ref{tab:Reveal-Metric} are proven to be either submodular, 
or approximatly submodular, and monotone increasing.
\cite{shamaiah2010greedy} provides proof for submodularity of {\em Max-logDet}.
The stronger property, modularity, holds for {\em Max-Trace} 
\cite{summers2016submodularity}.  
Though {\em Max-MinEigenValue} does not meet submodularity in 
general, it is recognized as approximately submodular \cite{carlone2017attention}.
Therefore, selecting row blocks 
(as well as the corresponding features) with these metrics can be 
approximated by greedy approach.

\subsection{Simulation of Good Feature Selection}
\begin{figure}[!htb]
\centering
\includegraphics[width=0.4\linewidth]{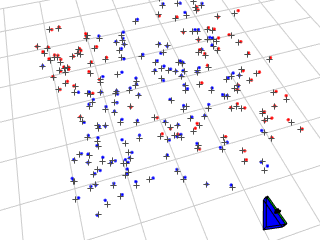} \hspace{15pt}
\includegraphics[width=0.4\linewidth]{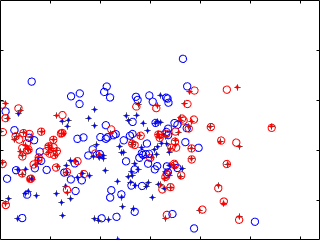}
\caption{Simulated pose optimization scenario.  Left: map view; 
  Right: camera view.  
  Selected features are red and unselected ones are blue. 
  \label{fig:Simu_PnP_Scene}} 	
\end{figure}

To identify the applicable cases of the good feature selection, and explore 
the matrix-revealing metrics that could guide good feature/row block subset 
selection, a simulation of least squares pose optimization was carried out.  
The simulation environment of \cite{LandmarkImpact}, which assumes
perfect data association, provides the testing framework.  The
evaluation scenario is depicted in Fig~\ref{fig:Simu_PnP_Scene}.  The
camera/robot is spawned at the origin of the world frame, and a fixed
number (e.g. 200 in this synthetic test) of 3D feature points are
randomly generated in front of the camera.  After applying a small
random pose transform to the robot/camera, the 2D projections of feature
points are measured and perfectly matched with known 3D feature points.
A Gauss-Newton optimizer estimates the random pose transform from the matches.

To simulate map error, the 3D feature points are perturbed with biased
noise (Gaussian with mean of 0.05m, and standard deviation of 0.05m).
The 2D measurements are also perturbed with two levels of measurement
error: zero-mean Gaussian with standard deviation of 1 and 2 pixel.
Subset size ranging from 80 to 200 are tested.  To be statistically
sound, 300 runs are repeated for each configuration.

Feature selection occurs prior to Gauss-Newton pose optimization, so
that only a subset of selected features is sent to the optimizer.
Each of the matrix-revealing metrics listed in Table~\ref{tab:Reveal-Metric} 
were tested.  
\begin{figure*}[!htb]
	\centering
	\includegraphics[width=0.24\linewidth]{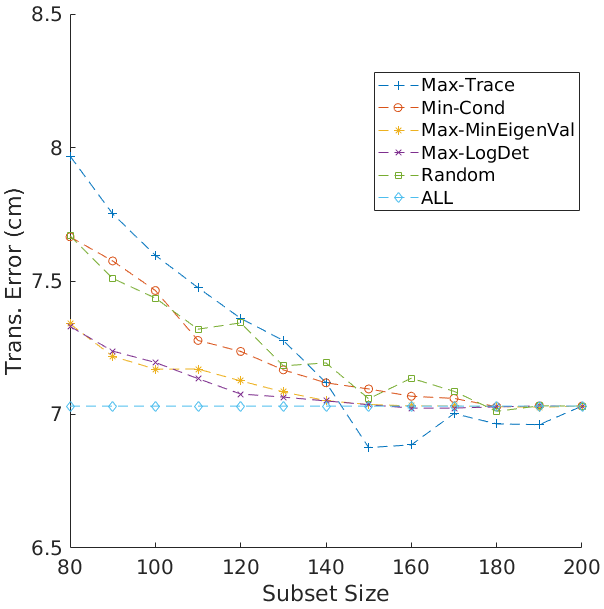}
	\includegraphics[width=0.24\linewidth]{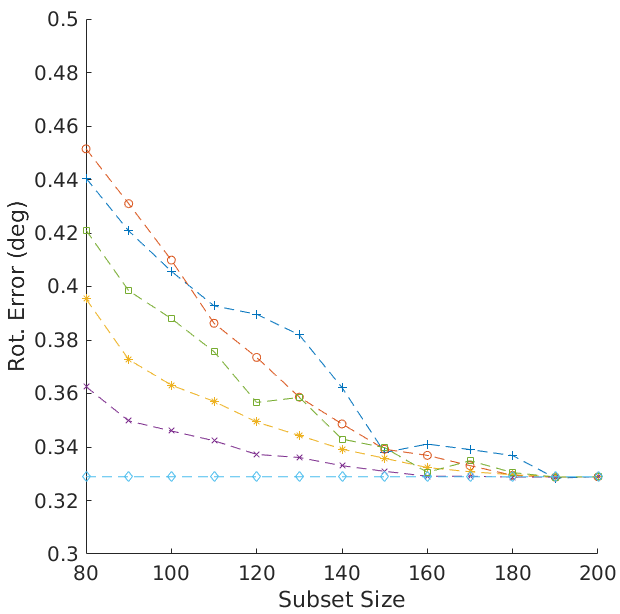}
	\hspace{5pt}
	\includegraphics[width=0.24\linewidth]{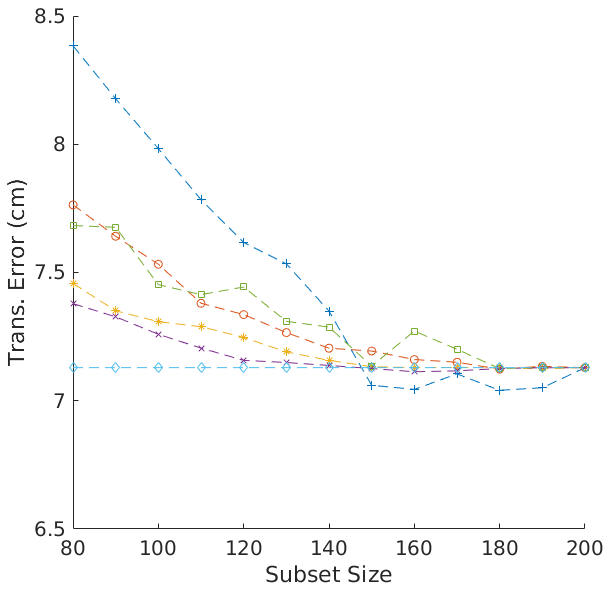}
	\includegraphics[width=0.24\linewidth]{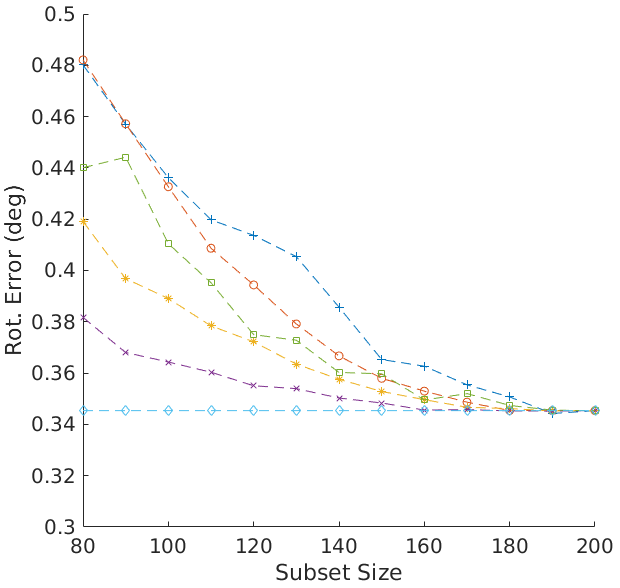}\\
  \vspace*{-1ex}
	\caption{Simulation results of least squares pose optimization.  
	First 2 columns: RMS of translational / rotational error under low noise.
	Last 2 columns: RMS of translational / rotational error under high noise.  \label{fig:Simu_PnP_Res}}
	\includegraphics[width=0.26\linewidth]{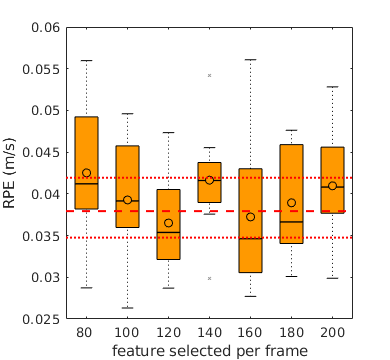}
	\hspace{5pt}
	\includegraphics[width=0.26\linewidth]{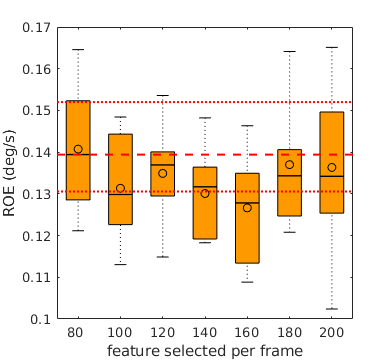}
	\hspace{5pt}
	\includegraphics[width=0.26\linewidth]{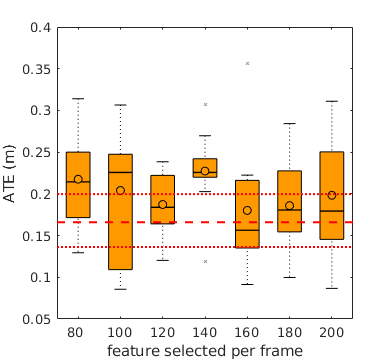}
	\\
	\caption{Tracking accuracy vs. number of good features selected per
    frame.  For {\em Max-logDet} ORB-SLAM, the error metric under each
    feature selection budget are presented as boxplot in orange;  for
    INL-ORB the mean (red dashed) and the 25-75 percentile (red
    dotted) lines are plotted.  \label{fig:Trend_MH_05}}
  \vspace*{-3ex}
\end{figure*}

Feature selection is done in three steps: 1) compute the full measurement 
Jacobian $H_x$ and projection Jacobian $H_p$, 2) combine the two into $H_c$, 
and 3) greedily select row blocks of $H_c(i)$, based on the matrix-revealing 
metric, until reaching the target subset size.  The simulation results 
are presented in Fig~\ref{fig:Simu_PnP_Res}.  
For reference, we also plot the simulation results with 
randomized subset selection ({\em Random}) and with all features available ({\em ALL}).

From Fig~\ref{fig:Simu_PnP_Res}, the {\em Max-logDet} metric has the
best overall performance.  Under both low and high level of 
residual noise, it more quickly approaches the baseline error ({\em
All}).  Though marginal, the translational error of {\em Max-logDet}
goes below the ALL baseline, while the rotational error equals
the baseline, once the subset size exceeds 160.  
The results point to the value of {\em Max-logDet} good
features selection.

\section{Efficient Max-logDet Subset Selection}

Subset selection with {\em Max-logDet} metric has been studied in fields
such as sensor selection \cite{shamaiah2010greedy} and feature selection
\cite{carlone2017attention}.  There, a simple greedy algorithm is
commonly used to approximate the original NP-hard combinational
optimization problem.  Since {\em Max-logDet} is submodular and monotone
increasing, the approximation ratio of the greedy approach is $1-1/e$
\cite{summers2016submodularity}, which is the best any polynomial time
algorithm can achieve under the assumption $P \neq NP$.

However, the computational cost of the greedy algorithm is too high 
for feature selection in real-time VO/VSLAM applications.
As reported in \cite{carlone2017attention} and confirmed by us, the time cost 
of greedy selection exceeds the real-time requirement (e.g. 30ms per
frame) with around 100 feature inputs.  To select $k$ feature out of $n$
candidates, the greedy algorithm has to run $k$ rounds. In each round it
considers all remaining candidates to identify the current best feature.  
Hence the total complexity of greedy algorithm is $\mathcal{O}(nk)$.  

To speed up the greedy feature selection, we explore the combination
of deterministic selection (e.g. the greedy algorithm) and randomized 
acceleration (e.g. random sampling).  One well-recognized method of combining
these two, is stochastic greedy \cite{mirzasoleiman2015lazier}.  
Each round of greedy selection evaluated a random subset of candidates
to identify the current ``best'' feature,  instead of going through all
$n$ candidates.  The random subset size $s$ is controlled by a decay
factor $\epsilon$: $s=\frac{n}{k}\log(\frac{1}{\epsilon})$.  
Complexity reduces to $\mathcal{O}(\log(\frac{1}{\epsilon})n)$.  

More importantly, the expected approximation guarantee of stochastic
greedy is proven to be $1-1/e-\epsilon$ \cite{mirzasoleiman2015lazier};
compare to $1-1/e$, the best approximation ratio of any polynomial time
algorithm \cite{summers2016submodularity}.  
Selecting a proper decay factor $\epsilon$ in stochastic greedy 
(e.g. $\epsilon=0.1$ in the following experiments), 
slightly lowers the optimum bound, while significantly speeding up
selection (16\% vs 43x).
Alg~\ref{alg:efficient_maxLogDet} summarizes the stochastic-greedy-based
{\em Max-logDet} feature selection algorithm.

\vspace*{-0.1in}
\begin{algorithm}[!htb]
\small
\KwData{ $H_c=\{H_c(1),\ H_c(2),\ ...\ ,\ H_c(n)\}$, $k$}
 \KwResult{ $H_c^{sub} \subseteq H_c,\ |H_c^{sub}|=k$}

 $H_c^{sub}\gets \emptyset$\;
 \While{$|H_c^{sub}|<k$}{
 	$H_c^R \gets \text{a random subset obtained by sampling }$ 
 	$s=\frac{n}{k}\log(\frac{1}{\epsilon})\text{ random elements from } H_c$\;
 	$H_c(i) \gets \arg\max_{H_c(i) \in H_c^R} \log \det (H_c(i)^T H_c(i)$
 	$+ [H_c^{sub}]^T [H_c^{sub}])$\;
	$H_c^{sub} \gets H_c^{sub}\cup H_c(i)$\;
 	$H_c \gets H_c \setminus H_c(i)$\;
 }
 \Return $H_c^{sub}$.
 \caption{\small Proposed efficient approximation algorithm for {\em Max-logDet} feature selection. \label{alg:efficient_maxLogDet}}
\end{algorithm}
\vspace*{-0.1in}

\section{Experimental Results on Real-time VSLAM}
This section evaluates the performance of the proposed {\em Max-logDet}
feature selection on a state-of-the-art feature-based monocular visual
SLAM system, ORB-SLAM \cite{ORBSLAM}.  By integrating the proposed
feature selection to the real-time tracking thread of ORB-SLAM,
we demonstrate significant improvement in pose tracking accuracy, while
the time cost of pose tracking only increases slightly.

Feature selection is done in the pose refinement function, $TrackLocalMap$, 
of the real-time tracking thread of ORB-SLAM.  All possible feature
matches found between the current frame and the local map are fed into
this function.  However, feature selection is not conducted on the whole
set of input matchings directly: the input set contains some outliers
(i.e. {\em non-inliers}), which affect the performance of pose
optimization when included.
Outlier rejection needs to be applied to the tracked features prior
to feature selection.  Due to the lack of explicit outlier rejection
in ORB-SLAM, we add an outlier rejection module by employing the
ORB-SLAM pose optimization code.  Pose optimization is conducted with
the whole set of feature matchings, then tracked features with high 
re-projection error are rejected.  Such an implementation of outlier
rejection is far from efficient, but it will kick out most of the
outliers.

Five feature selection approaches are implemented: 
1) {\em Quality}, which selects based on the ORB-matching score; 
2) {\em Bucket} \cite{geiger2011stereoscan}, which divides the frame
into grids and uniformly samples from them; 
3) {\em Observability} ({\em Obs}) \cite{zhang2015good}, which selects
based on observability over a short time (here, last 3 segments); 
4) {\em Max-logDet} ({\em MD}), which selects based on Alg~\ref{alg:efficient_maxLogDet}; and 
5) {\em Quality + MD}, which generates a subset of features based on the
ORB-matching score first, then selects from them using the {\em Max-logDet} 
algorithm.  
Two baseline approaches are included: 
1) {\em INL-ORB}, which has the explicit outlier rejection module on top of
original ORB-SLAM; and 
2) {\em ALL-ORB}, the original ORB-SLAM.  

Since the focus is on real-time pose tracking, all evaluations are
performed on the instantaneous output of pose tracking thread; key-frame
poses after posterior bundle adjustment are not used.
Relocalization and loop closing are disabled in all implementations.  
For ORB-SLAM with feature selection, the number of tracked
features used is fixed (100 features per frame).  
For the {\em Quality+MD} combination, a candidate pool of 200 features is
selected using Quality, from which the good feature subset is further
extracted based on the proposed {\em Max-logDet} algorithm.  Meanwhile,
for the baseline approaches {\em INL-SLAM} and {\em ALL-SLAM}, as many
as 2000 features can be used to optimize the pose per frame.

The benchmark used is the EuRoC MAV dataset~\cite{burri2016euroc}.  It
consists of stereo images and inertial data recorded from a micro aerial
vehicle.  Only the images from the left camera are used in this
monocular visual SLAM experiment.  In total 11 sequences are recorded
under 3 different indoor environments, with a total length of 19
minutes.  Challenging cases such as low-texture, illumination changes,
fast motion and motion blur are covered.  Each sequence has ground-truth 
from a motion capture system (Vicon or Leica MS50).

Due to the initialization procedure and multi-threaded structure of
ORB-SLAM, all approaches are run 10 times per sequence.  The platform
was an Intel i7 quadcore 4.20GHz CPU (passmark score of 2583 per thread) 
with ROS Indigo.  Accuracy of real-time pose tracking is evaluated with
three metrics \cite{sturm12iros_ws} between ground truth and SLAM
estimates (aligned to ground truth with a {\em Sim3} transform): 

\begin{table*}[t!]
	\small
	\centering
	\caption{Relative Position (m/s) / Orientation (deg/s) Error On EuRoC Sequences}
	\begin{tabular}{|c|c|c|c|c|c||c|c|}
		\hline 
		\textbf{ } & 
		\multicolumn{7}{c|}{\bfseries \small Approach} \\
		\textbf{\small Budget } & 
		\multicolumn{5}{c||}{\bfseries \small 100 Feature per Frame} &
		\multicolumn{2}{c|}{\bfseries \small 2000 Feature per Frame} \\
		\hline 
		\textbf{\small Seq.} & {\em Quality} & {\em Bucket} & {\em Obs} & {\em MD} & {\em Quality+MD} & {\em INL-ORB} & {\em ALL-ORB} \\
		\hline
		\textit{MH 01 easy} & 0.012 0.09 & 0.012 0.08 & 0.013 0.11 & \textbf{0.012 0.08} & 0.013 0.09 & 0.011 0.07 & \textbf{0.011 0.07} \\ 
		\textit{MH 02 easy} & 0.032 0.36 & 0.024 0.28 & 0.030 0.33 & \textbf{0.022 0.25} & 0.029 0.33 & 0.109 0.62 & \textbf{0.046 0.41} \\ 
		\textit{MH 03 med} & 0.027 0.12 & \textbf{0.026 0.12} & 0.029 0.14 & 0.053 0.27 & 0.028 0.12 & 0.028 0.11 & \textbf{0.027 0.10} \\ 
		\textit{MH 04 diff} & 0.096 0.28 & 0.077 0.25 & 0.084 0.26 & 0.077 0.31 & \textbf{0.064 0.19} & \textbf{0.071 0.20} & 0.094 0.26 \\  
		\textit{MH 05 diff} & 0.063 0.20 & 0.063 0.21 & 0.046 0.18 & 0.041 0.14 & \textbf{0.041 0.14} & \textbf{0.038 0.14} & 0.060 0.14 \\ 
		\textit{VR1 01 easy} & 0.038 0.46 & 0.038 0.45 & 0.038 0.47 & \textbf{0.038 0.45} & 0.038 0.45 & \textbf{0.038 0.45} & 0.038 0.45 \\ 
		\textit{VR2 01 easy} & 0.016 0.27 & 0.014 0.25 & 0.015 0.26 & 0.015 0.26 & \textbf{0.011 0.20} & \textbf{0.011} 0.24 & 0.012 \textbf{0.22} \\ 
		\textit{VR2 02 med} & 0.093 0.61 & 0.153 0.89 & 0.126 1.02  & 0.078 0.57 & \textbf{0.032 0.52} & \textbf{0.165} 0.87 & 0.200 \textbf{0.86} \\ 
		\hline	
		\textbf{\small Average} & 0.047 0.30   & 0.051 0.32   & 0.048 0.35   & 0.042 0.29   & \textbf{0.032 0.26}   & \textbf{0.059} 0.35   & 0.061 \textbf{0.32} \\
		\hline	
		\textbf{\small \# Seq. with Perf. Loss} & 6 & 6 & 6 & 5 & \textbf{3} & 5 & - \\
		\textbf{\small \# Seq. with Perf. Gain} & 2 & 2 & 2 & 3 & \textbf{5} & 3 & - \\
		\hline	
		\textbf{\small Average Perf. Loss} & 0.002 0.03   & 0.001 0.03   & 0.002 0.05   & 0.010 0.06   & {0.001 0.01}  & 0.032 0.05  & - \\
		\textbf{\small Average Perf. Gain} & {-0.060 -0.15} & -0.021 -0.07 & -0.028 -0.04 & -0.036 -0.15 & -0.047 -0.10 & -0.013 -0.02 & - \\
		\hline	
	\end{tabular} 
	\label{tab:EuRoC_Rel}
  \vspace*{-2ex}
\end{table*}

\noindent
1) \textbf{Absolute Trajectory Error (ATE)}, the
root-mean-square difference between the ground truth and the 
entire estimated trajectory; \\
2) \textbf{Relative Position Error (RPE)}, the average
drift of pose tracking over a short period of time; \\
3) \textbf{Relative Orientation Error (ROE)}, the average
orientation drift similar to RPE.   \\
RPE and ROE are averaging windows are 3 seconds.

\subsection{Accuracy vs. Subset Size}
The connection between the number of good features selected and the pose
optimization accuracy is assessed on one EuRoC sequence, \textit{MH 05 diff}.
Fast camera motion and changing lighting conditions challenge accurate
tracking and mapping.  When running on this sequence, the the
measurements and mapped features are expected to be noisy; good feature
selection should mitigate the effects of the noise.

Fig~\ref{fig:Trend_MH_05} consists of box plots for 10-runs of {\em
Max-logDet} ORB-SLAM on the example sequence under feature selection
budgets ranging from 80 to 200.  One plot for each evaluation metric. 
For reference, we plot (in red) the outcomes for inlier-only ORB-SLAM
(tracking up to 2000 features/frame).  
The improvement of feature selection is mostly significant on the
boxplot of ROE (between the budget of 100 and 180).  The improvement 
on RPE is less obvious: feature selection leads to a slight reduction of 
RPE for budgets of 120 and 160.  
The absolute metric (ATE) is less sensitive to subset selection.
In the subsequent evaluations on good feature selection, the smallest 
budget that leads to accuracy improvement will be used, 100 feature/frame. 

\subsection{Accuracy vs. Feature Selection Approaches}
Table~\ref{tab:EuRoC_Rel} summarizes the relative metrics (RPE and ROE). 
Each cell first reports the average RPE (units: m/s), then the average ROE
(units: deg/s).  
For each selection approach type (100 feat. and 2000 feat.), the lowest
relative errors per sequence are in bold.  
Three sequences are not included due to frequent failures (since 
relocalization is disabled).  

On almost all sequences, either the {\em MD} or the {\em Quality+MD} combination
has the lowest relative error of the feature selection approaches.
On challenging sequences such as \textit{MH 04 diff},
\textit{MH 05 diff} and \textit{VR2 02 med}, the combined approach
reduces the relative error significantly.  The exception is \textit{MH
03 med} where the combined approach results in a slightly higher RPE
than the lowest one (generated by {\em Bucket}).  Overall,
the {\em MD} approach reduces pose tracking error on several sequences
by exploiting the structural and motion information.  Integrating
{\em MD} with appearance information (i.e. {\em Quality}) further
improves performance. 

Now, compare {\em Quality+MD} with the two baselines.  
On sequences such as \textit{MH 02 easy}, \textit{MH 04 diff} and
\textit{VR2 02 med}, {\em Quality+MD} clearly leads to lower relative error.
Meanwhile on other sequences, the relative error of {\em Quality+MD}
is either the same as baselines or slightly worse.  The performance
gains on the harder sequences far outweigh the performance loss on the
easy sequences, as presented in the last 4 rows of Table~\ref{tab:EuRoC_Rel}.  
When under-performing, the {\em Quality+MD} approach has the lowest
performance loss.  When over-performing, it does so more often and by a
significant amount.  The average RPE and ROE scores for {\em Quality+MD}
improve by 47\% and 19\%, respectively, versus {\em ALL-ORB}.

\subsection{Good Feature ORB-SLAM vs. Other VO/VSLAM}
The accuracy improvement of good feature selection using Quality+MD
is further demonstrated by comparing against other state-of-the-art
VO/VSLAM methods.  Two direct approaches, SVO \cite{SVO2017} and
DSO \cite{DSO2017}, are chosen as baselines.  For fair comparison, both 
SVO and DSO are evaluated under the same configuration as above: 
1) monocular vision input only with real-time enforcement, 
2) up to 2000 (patch) matchings per frame, 
3) real-time pose tracking results of the entire sequence 
being evaluated (both \cite{SVO2017} and \cite{DSO2017} remove the 
beginning part with strong motion in evaluation), and 
4) only those succeeding for all 10 runs are reported (no tracking
failure allowed).  
Performance is measured with absolute translation error (ATE), 
as per \cite{SVO2017}.  
Table~\ref{tab:EuRoC_Abs} reports the ATEs.

With {\em Quality+MD} feature selection, the ATE on sequences \textit{MH 02
easy}, \textit{MH 04 diff}, and \textit{VR2 02 med} are significantly 
reduced, while the accuracy advantage are preserved on the rest.  
The error metrics statistics given in the last three rows indicate that
{\em Quality+MD} ORB-SLAM has the lowest average ATE,
as well as the lowest maximum ATE compared to the approaches evaluated.  
The two direct baselines do not perform as well: SVO has the worst 
ATE on all 10 trackable sequences; DSO only tracks on 5 sequences 
completely, and has the 2nd worst average ATE.

\begin{table}[tb!]
	\small
	\centering
	\caption{Absolute Translation Error (m) On EuRoC Sequences}
	\begin{tabular}{|c|c|c|c||c||c|}
		\hline 
		\textbf{ } & 
		\multicolumn{5}{c|}{\bfseries \small Approach} \\
		\textbf{ } & 
		\multicolumn{3}{c||}{\bfseries ORB-SLAM} &
		\bfseries SVO & \bfseries DSO \\
		\textbf{\small Seq.} & {\em ALL} & {\em INL} & {\em Quality+MD} &   &   \\
		\hline
		\textit{MH 01 easy}  & \textbf{0.03} & \textbf{0.03} & 0.04 & 0.30 & - \\ 
		\textit{MH 02 easy}  & 0.33 & 0.70 & \textbf{0.15} & - & - \\ 
		\textit{MH 03 med}   & \textbf{0.04} & 0.05 & 0.05 & 0.39 & 0.75 \\ 
		\textit{MH 04 diff}  & 0.96 & 0.48 & \textbf{0.41} & 5.82 & - \\  
		\textit{MH 05 diff}  & 0.27 & \textbf{0.17} & 0.22 & 4.15 & - \\ 
		\textit{VR1 01 easy} & 0.04 & 0.04 & 0.04 & 0.77 & 0.64 \\ 
		\textit{VR1 02 med}  & - & - & - & 0.77 & 0.53 \\ 
		\textit{VR1 03 diff} & - & - & - & 0.65 & - \\ 
		\textit{VR2 01 easy} & 0.04 & 0.04 & 0.04 & 0.18 & 0.29 \\ 
		\textit{VR2 02 med}  & 0.75 & 0.61 & \textbf{0.12} & 1.57 & 1.04 \\ 
		\textit{VR2 03 diff} & - & - & - & 1.66 & - \\ 
		\hline	
		\textbf{\small Average} & 0.31 & 0.27 & \textbf{0.13} & 1.63 & 0.65 \\
		\textbf{\small Min.} & \textbf{0.03} & \textbf{0.03} & 0.04 & 0.18 & 0.29 \\
		\textbf{\small Max.} & 0.96 & 0.70 & \textbf{0.41} & 5.82 & 1.04 \\
		\hline	
	\end{tabular} 
	\label{tab:EuRoC_Abs}
    \vspace*{0.05in}
	\caption{Time Cost (ms) Per Frame Breaking Down For Pose Tracking}
	\begin{tabular}{|c|c|c|c||c|}
		\hline 
		\textbf{ } & 
		\multirow{2}{*}{\twoRowCell{ \textbf{\small Base } }{ }} & 
		\multirow{2}{*}{\twoRowCell{ \textbf{\small Feat. } }{ \textbf{\small Sel. } }} & 
		\multirow{2}{*}{\twoRowCell{ \textbf{\small Pose } }{ \textbf{\small Opt. } }} & 
		\multirow{2}{*}{\twoRowCell{ \textbf{\small Total } }{ }} \\
		&  &  &  &  \\		
		\hline 
		{\em Quality} 	& 28.6 & 0.00 & 0.35 & 29.0 (-0.7\%) \\
		{\em Bucket} 	& 28.5 & 0.11 & 0.34 & 29.0 (-0.7\%) \\
		{\em Obs} 		& 28.5 & 2.6  & 0.32 & 31.4 (+7.5\%) \\
		{\em MD} 		& 28.6 & 3.0  & 0.40 & 32.0 (+9.6\%) \\
		{\em Quality+MD}& 28.5 & 1.5  & 0.34 & 30.3 (+3.8\%) \\
		\hline 
		{\em INL-ORB} 	& 28.8 &  -   & 2.4 & 31.3 (+7.2\%) \\
		{\em ALL-ORB} 	& 26.5 &  -   & 2.6 & 29.2 \\
		\hline	
	\end{tabular} 
	\label{tab:EuRoC_Time}
  \vspace*{-0.1in}
\end{table}

\subsection{Efficiency vs. Feature Selection Approaches}
Table~\ref{tab:EuRoC_Time} present a breakdown of the computation time
for each feature selection approach (averaged over all EuRoC sequences).  
The {\em Base} column measures the pre-processing steps (ORB extraction,
initial tracking, and outlier rejection) before feature selection.
Due to the outlier rejection step, all methods except ALL-SLAM, incur
increased timing.
Of the structural-based selection approaches, {\em Quality+MD} is the 
2nd fastest.  {\em Bucket} is extremely efficient, but does not improve
as much the accuracy.  
When comparing {\em Quality+MD} to baseline
ORB-SLAM, outlier rejection time cost is almost offset by the time
savings in pose optimization.  We imagine better implemented outlier
rejection or integration of outlier rejection and feature
(inlier) selection, could consume less time than {\em ALL-ORB}, while
still enhancing performance.

\section{Conclusion}
This paper presented the idea of good feature selection for least
squares pose optimization.  Under a biased noise assumption, selecting a
subset of features should improve optimization accuracy.
The connection between matrix subset selection methods and the solution
conditioning of least squares optimization was discussed.  
Through a controlled experiment, the {\em Max-logDet} matrix revealing
metric was shown to perform best.  
For rapid subset selection, a near optimal heuristic approach to 
{\em Max-logDet} is used.
Integrating the proposed good feature selection approach with a feature point
quality scoring selector and outlier rejection leads to a more accurate
visual odometry within a SLAM system with nearly the same computational
cost.

\bibliographystyle{IEEEtran}
\bibliography{draft}

\end{document}